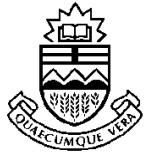

# University of Alberta

# A New Paradigm for Minimax Search

by

Aske Plaat, Jonathan Schaeffer, Wim Pijls and Arie de Bruin

# A New Paradigm for Minimax Search

Aske Plaat, Erasmus University, *plaat@theory.lcs.mit.edu*
Jonathan Schaeffer, University of Alberta, *jonathan@cs.ualberta.ca*
Wim Pijls, Erasmus University, *whlmp@cs.few.eur.nl*
Arie de Bruin, Erasmus University, *arie@cs.few.eur.nl*

| | |
|---|---|
| Erasmus University, | University of Alberta, |
| Department of Computer Science, | Department of Computing Science, |
| P.O. Box 1738, | 615 General Services Building, |
| 3000 DR Rotterdam, | Edmonton, Alberta, |
| The Netherlands | Canada T6G 2H1 |

**Abstract**

This paper introduces a new paradigm for minimax game-tree search algorithms. MT is a memory-enhanced version of Pearl's *Test* procedure. By changing the way MT is called, a number of best-first game-tree search algorithms can be simply and elegantly constructed (including SSS*).

Most of the assessments of minimax search algorithms have been based on simulations. However, these simulations generally do not address two of the key ingredients of high performance game-playing programs: iterative deepening and memory usage. This paper presents experimental data from three game-playing programs (checkers, Othello and chess), covering the range from low to high branching factor. The improved move ordering due to iterative deepening and memory usage results in significantly different results from those portrayed in the literature. Whereas some simulations show Alpha-Beta expanding almost 100% more leaf nodes than other algorithms [12], our results showed variations of less than 20%.

One new instance of our framework (MTD-f) out-performs our best alpha-beta searcher (aspiration NegaScout) on leaf nodes, total nodes and execution time. To our knowledge, these are the first reported results that compare both depth-first and best-first algorithms given the same amount of memory.

**Contents**

# 1 Introduction

For over 30 years, Alpha-Beta has been the algorithm of choice for searching game trees. Using a simple left-to-right depth-first traversal, it is able to efficiently search trees [9, 15]. Several important enhancements were added to the basic Alpha-Beta framework, including iterative deepening, transposition tables [25], the history heuristic [23] and minimal window searching [2, 14, 19]. Several studies showed that the algorithm's efficiency was approaching optimal and that there was little room for improvement [5, 6, 23].

In 1979, Stockman introduced the SSS* algorithm which was able to provably expand fewer leaf nodes than Alpha-Beta by adopting a best-first search strategy [26]. Scientific skepticism quickly followed as it became evident that SSS* had serious implementation drawbacks. These problems revolved around the OPEN list, a data structure whose size was exponential in the depth of the search tree, and the expensive operations required to maintain the list in sorted order [21]. Nevertheless, the potential for building smaller search trees had been demonstrated. Unfortunately, it seemed that SSS* was an interesting idea for the theoreticians, but it failed to become an algorithm used by the practitioners.

This paper introduces the *Memory-enhanced Test* (MT) algorithm, a variation on Pearl's Test procedure [14]. This routine efficiently searches a game tree to answer a binary question. MT can be called from a simple driver routine which may make repeated calls to MT. By using memory to store previously seen nodes (a standard transposition table), MT can be used efficiently to re-search trees.

A search yielding a single binary decision is usually not useful for game-playing programs. Repeated calls to MT can be used to home in on the minimax value. By constructing different driver programs that call MT, different algorithms can be created. In particular, a variety of best-first algorithms can be implemented using depth-first search, such as SSS* and DUAL* [12, 20]. The surprising result is that a series of binary-valued searches is more effective at determining the minimax value than searches over a wide range of values. It is interesting to note that the basic Alpha-Beta algorithm *cannot* be created using MT, because its wide search window causes fewer cutoffs than MT-based algorithms.

The MT drivers (MTD) can easily be changed to create new algorithms. Some new algorithms are introduced in this paper and one of them, MTD-f, out-performs SSS*, DUAL*, Alpha-Beta and PVS/NegaScout [2, 19] on average. While SSS* uses $+\infty$ as its initial search bound and DUAL* uses $-\infty$, MTD-f uses the result of the previous iteration in an iterative deepening search. Starting with a bound closer to the expected outcome increases search efficiency, as we will show.

Most papers in the literature compare game-tree search algorithms using simulations. Typically, the simulations do not use iterative deepening or transposition tables (for example, [8, 12, 13]). Since these are necessary ingredients for achieving high performance in real applications, this is a serious omission. Rather than use simulations, our results have been taken from three game-playing programs. These include the checkers program Chinook (a game with low branching factor), the Othello program Keyano (medium branching factor) and the chess program Phoenix (high branching factor). All three programs are well-known in their respective domains. Experimentally comparing a variety of best-first and depth-first algorithms using the same storage requirements paints a very different picture of the relative strengths of the algorithms. In particular, the results of non-iterative deepening experiments reported in the literature are misleading. With iterative deepening and transposition tables, move ordering is improved to the point where all the algorithms start approaching the minimal search tree and the differences between the algorithms become less pronounced.

Since binary decision procedures are shown to be more effective than wide-windowed searches, there is no good reason for continued usage of wide-windowed Alpha-Beta. Although every introductory book on artificial intelligence discusses Alpha-Beta, we would argue that MT is a simpler algorithm, is easier to understand, and is more efficient than Alpha-Beta. To put it bluntly, why would anyone want to use Alpha-Beta?

## 2 Memory-enhanced Test

This section discusses the advantages of using a binary decision procedure for doing game-tree search.

### 2.1 A Binary Decision Procedure

Pearl introduced the concept of a proof procedure for game trees in his Scout algorithm [16]. A proof procedure takes a search assertion (for example, the minimax value of the tree is $\geq 10$) and returns a binary value (assertion is true or false). It was invented to simplify the analysis of the efficiency of Alpha-Beta. In doing the analysis, Pearl discovered that an algorithm using a binary decision procedure could search trees more efficiently. The proof procedure was called Test, since it was used to test whether the minimax value of a subtree would lie above or below a specified threshold. Scout was the driver program that called Test. Enhancements to the algorithm were developed and analyzed (this is well documented in [20]). It turns out that Scout-based algorithms will never consider more unique leaf nodes than would Alpha-Beta. For the special case of a perfectly ordered tree both Alpha-Beta and the Scout-variants search the so-called *minimal search tree* [9]. Simulations have shown that, on average, Scout-variants (such as NegaScout) significantly out-perform Alpha-Beta [8, 12]. However, when used in practice with iterative deepening, aspiration searching and transposition tables, the quality of the move ordering greatly increases. As a result, the relative advantage of NegaScout significantly decreases [23].

Test does an optimal job of answering a binary-valued question. However, game-tree searches are usually over a range of values. We would like to use the efficiency of Test to find the minimax value of a search tree. Repeated calls to Test will be inefficient, unless Test is modified to reuse the results of previous searches. Enhancing Test with memory yields a new algorithm which we call MT (short for *Memory-enhanced Test*), as illustrated in figure 1. The storage can be organized as a familiar transposition table [10]. Before a node is expanded in a search, a check is made to see if the value of the node is available in memory, the result of a previous search. Later on we will see that adding storage has some other benefits that are crucial for the algorithm's efficiency.

We would like to formulate our version of Test as clear and concise as possible, using the often used Negamax formulation [9]. Here we encounter a problem. Suppose we want to test whether the minimax value of maximizing node $n$ is at least $\omega$. A call Test$(n, \omega)$ would do this. A child returning a value $\geq \omega$ is sufficient to answer the question, and therefore causes a cutoff. Unfortunately, for a minimizing node, a cutoff should occur only when its return value $-g > -\omega$ (using a Negamax formulation). There is a simple way of removing the ">"/"≥" asymmetry: This can be obtained by using input parameters of the form $\omega - \varepsilon$ and $\omega + \varepsilon$, where $\varepsilon$ is a value smaller than the difference between any two leaf node evaluations. Then MT$(n, \omega - \varepsilon)$ guarantees that all nodes in the tree are either greater than or less than $\omega - \varepsilon$; no ties are possible. So, there are two differences between MT and Test. One is that MT is called with $\omega - \varepsilon$ instead of $\omega$, so that there is no need for the "=" part in the cutoff test, obviating extra tests in the code of MT or artificial incrementing/decrementing of the input parameter (as in [11]). The second (and more fundamental) difference is that MT uses storage to pass on search results from one pass to the next, making efficient multi-pass searches possible.

Figure 1 shows the pseudo-code for MT. The routine assumes an *evaluate* routine that assigns a value to a node. Determining when to call *evaluate* is application-dependent and is hidden in the definition of the condition $n = leaf$. For a depth $d$ fixed-depth search, a leaf is any node that is $d$ moves from the root of the tree. The search returns an upper or lower bound on the search value at each node, denoted by $f^+$ and $f^-$ respectively. Before searching a node, the transposition table information is *retrieved* and, if it has been previously searched deep enough, the search is cutoff. At the completion of a node, the bound on the value is *stored* in the transposition table. The bounds stored with each node are denoted using Pascal's *dot*-notation.

```
function MT(n, γ) → g;
{ precondition: γ ≠ any leaf-evaluation; MT must be called
  with γ = ω − ε to prove g < ω or g ≥ ω }
    if retrieve(n) = found then
        if n.f⁻ > γ then return n.f⁻;
        if n.f⁺ < γ then return n.f⁺;
    if n = leaf then
        n.f⁺ := n.f⁻ := g := evaluate(n);
    else
        g := −∞;
        c := firstchild(n);
        while g < γ and c ≠ ⊥ do
            g := max(g, −MT(c, −γ));
            c := nextbrother(c);
        if g < γ then n.f⁺ := g else n.f⁻ := g;
    store(n);
    return g;
```

Figure 1: MT, a memory-enhanced version of Pearl's Test
.

Usually we want to know more than just a bound on the minimax value. Using repeated calls to MT, the search can home in on the minimax value until it is found. To achieve this, MT must be called from a driver routine. One idea for such a driver would

```
function MTD+∞(n) → f;
    g := +∞;
    repeat
        bound := g;
        g := MT(n, bound − ε);
    until g = bound;
    return g;
```

Figure 2: MTD+∞: A sequence of MT searches to find $f$.

be to start at an upper bound for the search value, $f^+ = +\infty$. Subsequent calls to MT can lower this bound until the minimax value is reached. Figure 2 shows the pseudocode for this driver called MTD+∞. The variable $g$ is at all times an upper bound $f^+$ on the minimax value of the root of the game tree [18]. Surprisingly, MTD+∞ expands the same leaf nodes in the same order as SSS*, provided that no collisions occur in the transposition table, and that its size is of order $O(w^{\lceil d/2 \rceil})$ [4, 18].

*2.2 Example*

The following example illustrates how MTD+∞ traverses a tree as it computes a sequence of upper bounds on the game value. For ease of comparison, the example tree in figure 3 is the same as has been used in other papers [16]. Figures 4–7 show the four stages involved in building the tree. In these figures, the $g$ values returned by MT are given beside each interior node. Maximizing nodes are denoted by a square; minimizing nodes by a circle. The discussion uses the code in figures 1 and 2.

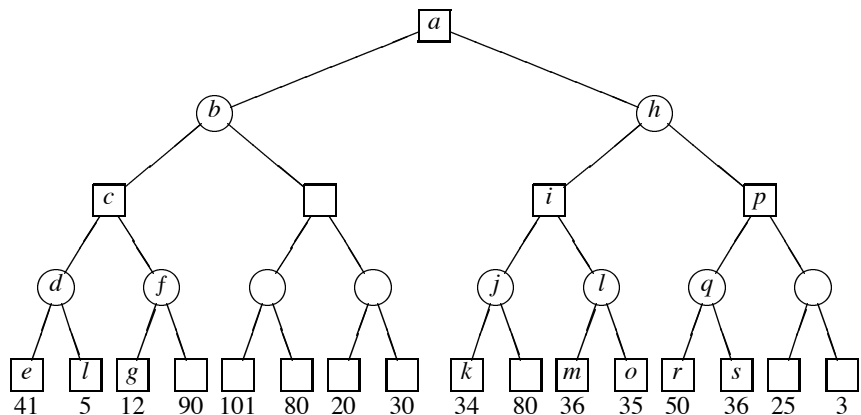

Figure 3: Example tree for MTD+∞.

*First Pass:* (figure 4)
MTD+∞ initially starts with a value of +∞ (1000 for our purposes). MT expands all branches at max nodes and a single branch at a min node. In other words, the maximum number of cutoffs occur since all values in the tree are $< 1000 − \varepsilon$ ($< \infty$). Figure 4 shows the resulting tree that is searched and stored in memory. The minimax value of this tree is 41. Note that at max nodes $a, c$ and $i$ an $f^+$ value is saved in the transposition

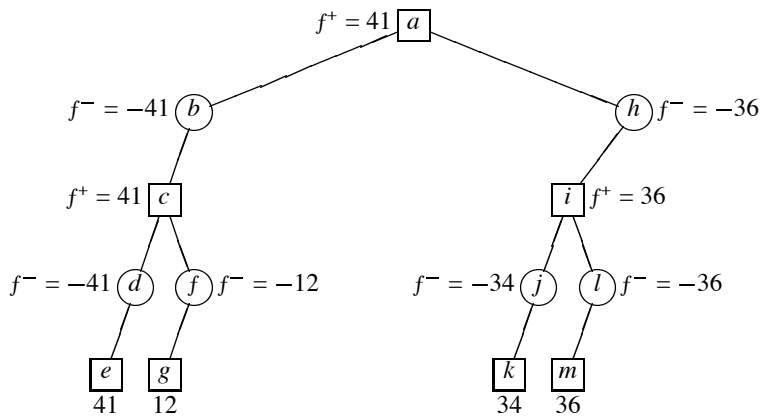

Figure 4: Pass 1.

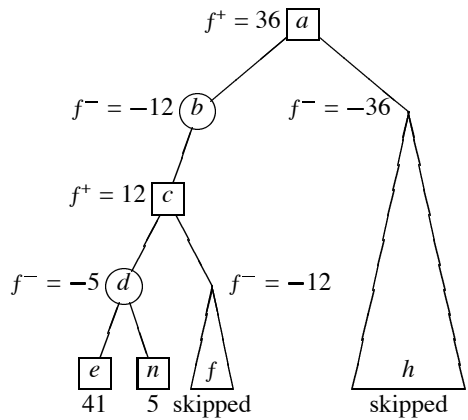

Figure 5: Pass 2.

table, while an $f^-$ value is saved at min nodes $b, d, f, h, j$ and $l$. Both bounds are stored at leaf nodes $e, g, k$ and $m$ since the minimax value for that node is exactly known. These stored nodes will be used in the subsequent passes.

*Second Pass:* (see figure 5)
MT is called again since the previous call to MT improved the upper bound on the minimax value from 1000 to 41 (the variable $g$ in figure 2). MT now attempts to find whether the tree value is less than or greater than $41 - \varepsilon$. The left-most path from the root to a leaf, where each node along that path has the same $g$ value as the root, is called the *critical path* or *principal variation*. The path $a, b, c, d$ down to $e$ is the critical path that is descended in the second pass. $e$ does not have to be reevaluated; its value comes from the transposition table. Since $d$ is a minimizing node, the first child $e$ does not cause a cutoff (value $< -(41 - \varepsilon)$) and child $n$ must be expanded. $n$'s value gets backed up to $c$, who then has to investigate child $f$. The bound on $f$, computed in the previous pass, causes the search to stop exploring this branch immediately. $c$ takes on the maximum of 12 and 5, and this becomes the new value for $b$. Since $h$ has a value $< 41 - \varepsilon$ (from the previous pass), the search is complete; both of $a$'s children prove that $a$ has a value less than 41. The resulting tree defines a new upper bound of 36 on the minimax value.

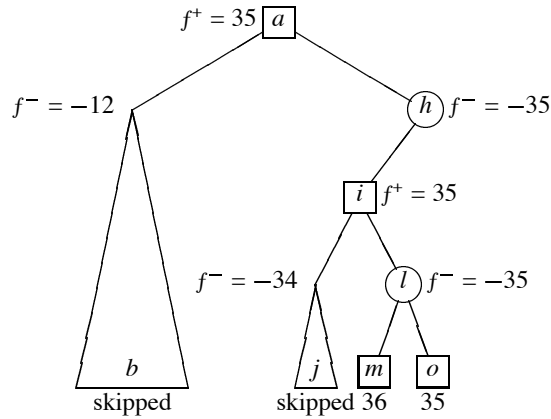

Figure 6: Pass 3.

*Third Pass:* (see figure 6)
The search now attempts to lower the minimax value below 36. From the previous search, we know $b$ has a value $< 36$ but $h$ does not. The algorithm follows the critical path to the node giving the 36 ($h$ to $i$ to $l$ to $m$). $m$'s value is known, and it is not less than 36. Thus node $o$ must be examined, yielding a value of 35. From $i$'s point of view, the new value for $l$ (35) and the bound on $j$'s value (34 from the first pass) are both less than 36. $i$ gets the maximum (35) and this gets propagated to the root of the tree. The bound on the minimax value at the root has been improved from 36 to 35.

*Fourth Pass:* (see figure 7)
The previous search lowered the bound from 36 to 35, meaning convergence has not occurred and the search must continue. The search follows the critical path $a, h, i, l$ and $o$. At node $l$, both its children immediately return without having been evaluated; their value is retrieved from the transposition table. Note that the previous pass stored an $f^-$ value for $l$, while this pass will store a $f^+$. There is room for optimization here by recognizing that all of *l*'s children have been evaluated and thus we know the exact value for *l* (see [17, 19]). The value of *l* does not change and *j*'s bound precludes it from being searched, thus *i*'s value remains unchanged. *i* cannot lower *h*'s value (no cutoff occurs), so the search explores *p*. *p* considers *q* which, in turn, must search *r* and *s*. Since *p* is a maximizing node, the value of *q* (36) causes a cutoff. Both of *h*'s children are greater than $35 - \varepsilon$. Node *a* was searched attempting to show whether it's value was less than or greater than $35 - \varepsilon$. *h* provides the answer; greater than. This call to MT *fails high*, meaning we have a lower bound of 35 on the search. The previous call to MT established an upper bound of 35. Thus the minimax value of the tree is proven to be 35.

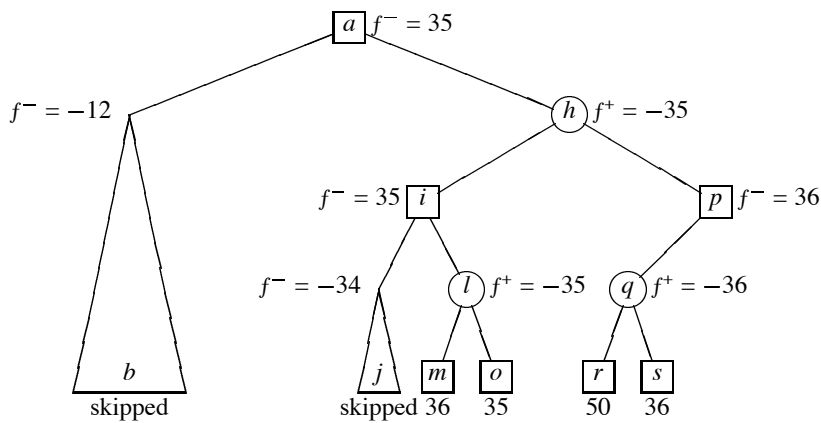

Figure 7: Pass 4.

## *2.3 Discussion*

MTD+∞ causes MT to expand the same leaf nodes in the same order as SSS* (a proof can be found in [18]). How does MT, a depth-first search procedure, examine nodes in a best-first manner? The value of $f^+ - \varepsilon$ causes MT to explore only nodes that can lower the upper bound at the root. This is the best-first expansion order of SSS*: in the third pass of the example, lowering the value of node *b* cannot influence $f(a)$, but node *h* can. Selecting node *h* gives a best-first expansion order.

The example illustrates that storage is critical to the performance of a multi-pass MT algorithm. Without it, the program would revisit interior nodes without the benefit of information gained from the previous search. Instead, MT can retrieve valuable search information for a node, such as an upper or lower bound, using a relatively cheap table lookup. The storage table provides two benefits: (1) preventing unnecessary node re-expansion, and (2) guiding MT along the critical path, ensuring a best-first selection scheme. Both are necessary for the efficiency of the algorithm.

One could ask the question whether a simple one-pass Alpha-Beta search would not be as efficient. To see why MT is more efficient it helps to think of MT as a null-window Alpha-Beta search ($\beta = \alpha + 1$) with a transposition table. Null-window searching has been shown equivalent to Test and has been discussed by numerous authors [2, 12, 16, 20]. Various papers point out that a tighter Alpha-Beta window causes more cutoffs than a wider window, all other things being equal (for example, [2, 18]). Since MT does not re-expand nodes from a previous pass, it cannot have fewer cutoffs than wide-windowed Alpha-Beta for new leaf nodes. This implies that any sequence of MT calls will be more efficient (it will never expand more leaf nodes and usually significantly less nodes) than a call to Alpha-Beta with window $(-\infty, +\infty)$.

## 3 Drivers for MT

Having seen one driver for MT, the ideas can be encompassed in a generalized driver routine. The driver can be regarded as providing a series of calls to MT to successively refine bounds on the minimax value.

The driver code can be parameterized so that one piece of code can construct a variety of algorithms. The two parameters needed are:

***first*** The *first* starting bound for MT.

***next*** A search has been completed. Use its result to determine the *next* bound for MT.

```
function MTD(first, next, n) → f;
    f⁺ := +∞; f⁻ := −∞;
    bound := g := first;
    repeat
        { The next operator must set the variable bound }
        next;
        g := MT(n, bound − ε);
        if g < bound then f⁺ := g else f⁻ := g;
    until f⁻ = f⁺;
    return g;
```

Figure 8: A framework for MT drivers.

Using these parameters, an algorithm using our MT driver, MTD, can be expressed as *MTD(first, next)*. The corresponding pseudocode can be found in figure 8. A number of interesting algorithms can easily be constructed using MTD. In the following, $+\infty$ and $-\infty$ are used as the upper and lower bounds on the range of leaf values. In actual implementations, these bounds are suitably large finite numbers.

- SSS* [MTD+∞]
  MTD+∞ is just SSS* and can be described as:

  $$\text{MTD}(+\infty, bound := g).$$

- DUAL* [MTD−∞]
  The literature describes a dual version of SSS*, where minimization is replaced by maximization, the OPEN list is kept in reversed order, and the start value is $-\infty$ [12, 20]. This algorithm becomes:

  $$\text{MTD}(-\infty, bound := g + 1).$$

  The advantage of DUAL* over SSS* lies in the search of odd-depth search trees. The trees that define such a bound are called *solution trees* [17, 26]. An upper bound is defined by a max solution tree, a lower bound by a min solution tree. In a max solution tree all children of max nodes are incorporated, and exactly one child of a min node. Dually for a min solution tree.

  Since SSS* builds and refines a max solution tree of size $O(w^{\lceil d/2 \rceil})$ on uniform trees, DUAL* builds and refines min solution trees of size $O(w^{\lfloor d/2 \rfloor})$ [20]. Since the min solution trees are smaller, fewer leaf nodes are expanded and less storage is required. If, for example, $w = 40$ and $d = 9$ then it makes a big difference whether one has to search and store a tree with $40^4$ or $40^5$ leaves.

  (Alternatively, for reasons of symmetry, we could replace the seventh line of figure 8 by $g := \text{MT}(n, bound + \varepsilon)$. In this way DUAL* can be expressed as $\text{MTD}(-\infty, bound := g)$, which is more like the driver for SSS*.)

- MTD-bi
  Since MT can be used to search from above (SSS*) as well as from below

(DUAL*), an obvious try is to bisect the interval and start in the middle. Since each pass produces an upper or lower bound, we can take some pivot value in between as the next center for our search. In MTD terms, bisecting the range of interest becomes:

$$\text{MTD}(\text{avg}(+\infty, -\infty), \text{avg}(f^+, f^-))$$

where *avg* denotes the average of two values. To reduce big swings in the pivot value, some kind of aspiration searching will be beneficial in many application domains [23].

Coplan introduced an equivalent algorithm which he named *C\** [3]. He does not state the link with best-first SSS*-like behavior, but does prove that C* dominates Alpha-Beta in the number of leaf nodes evaluated, provided there is enough storage. (Weill presents a NegaMax formulation of C* in [27].)

- MTD-f
  Rather than arbitrarily using the mid-point as an approximation, any information on where the value is likely to lie can be used as a better approximation. Given that iterative deepening is used in many application domains, the obvious approximation for the minimax value is the result of the previous iteration. In MTD terms this algorithm becomes:

  $$\text{MTD}(\text{approximation}, \textbf{if } g < bound \textbf{ then } bound := g \textbf{ else } bound := g + 1).$$

  MTD-f can be viewed as starting close to $f$, and then doing either SSS* or DUAL*, skipping a large part of their search path.

- MTD-step
  Instead of making tiny jumps from one bound to the next, as in all the above algorithms except MTD-bi, we could make bigger jumps:

  $$\text{MTD}(+\infty, bound := \max(f^-_{root} + 1, g - stepsize))$$

  (or the dual version) where *stepsize* is some suitably large value.

- MTD-best
  If we are not interested in the game value itself, but only in the best move, then a stop criterion suggested by Berliner can be used [1]. Whenever the lower bound of one move is not lower than the upper bounds of all other moves, it is certain that this must be the best move. To prove this, we have to do less work than when we solve for $f$, since no upper bound on the value of the best move has to be computed. We can use either a prove-best strategy (establish a lower bound on one move and then try to create an upper bound on the others) or disprove-rest (establish an upper bound on all moves thought to be inferior and try to find a lower bound on the remaining move). The stop criterion in figure 8 must be changed to $f^-_{bestmove} \geq f^+_{othermoves}$. Note that this strategy has the potential to build search trees *smaller* than the minimal search tree.

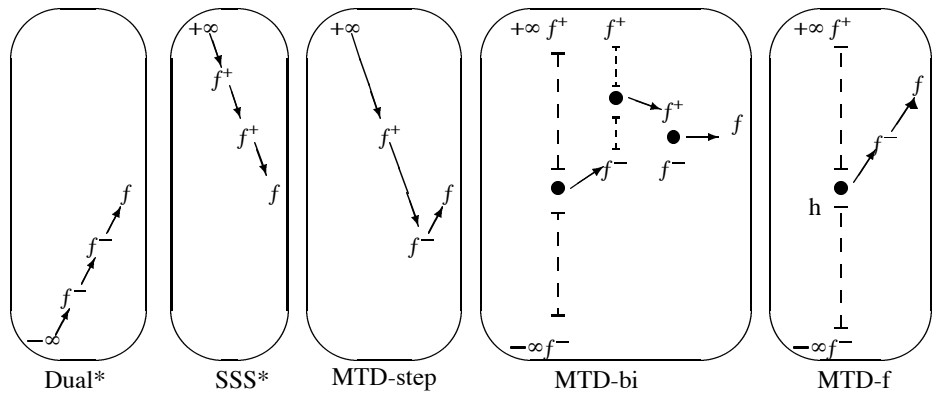

Figure 9: MT-based algorithms.

The knowledge which move should be regarded as best, and a suitable value for a first guess, might be obtained from a previous iteration in an iterative deepening scheme. The notion of which move is best might change during the search. This makes for a slightly more complicated implementation.

Note that while all the above algorithms use a transposition table, not all of them need to save both $f^+$ and $f^-$ values. For example, since SSS* always tries to lower the bound at the root, the transposition table needs to store only one bound.

Figure 9 illustrates the different strategies used by the above algorithms for converging on the minimax value.

The MTD framework has a number of important advantages for reasoning about game-tree search algorithms.

## 3.1 Elegance

Formulating a seemingly diverse collection of algorithms into one unifying framework focuses attention on the fundamental differences. For example, the framework allows the reader to see just how similar SSS* and DUAL* really are, and that these are just special cases of calling Test. The drivers concisely capture the algorithm differences. MTD offers us a high-level paradigm that facilitates the reasoning about important issues like algorithm efficiency and memory usage, without the need for low-level details.

## 3.2 Efficiency

All the algorithms presented are based on MT. Since MT is equivalent to a null-window Alpha-Beta call (plus storage), they search less nodes than the inferior one-pass Alpha-Beta($-\infty, +\infty$) algorithm (see also [18]).

## 3.3 Memory Usage

An important issue concerning the efficiency of MT-based algorithms is memory usage. SSS* can be regarded as manipulating one max solution tree in place [18]. A max solution tree has only one successor at each min node and all successors at max nodes, while the converse is true for min solution trees. Whenever the upper bound is lowered, a new (better) subtree has been expanded. The subtree rooted in the left brother of the new subtree can be deleted from memory, since it is inferior [12, 22]. Since the new brother is better, the previous brother nodes will never contain the critical path again and will never be explored (other than as transpositions). For a max solution tree, this implies that there is always only *one* live child at a min node; better moves just replace it. Given a branching factor of $w$ and a tree of depth $d$, the space complexity of a driver causing MT to construct and refine one max solution tree is therefore of the order $O(w^{\lceil d/2 \rceil})$, and a driver manipulating one min solution tree is order $O(w^{\lfloor d/2 \rfloor})$. Some of the algorithms, such as MTD-bi, use two bounds. These require a max and a min solution tree, implying that the storage requirements are $O(w^{\lceil d/2 \rceil})$. A simple calculation and empirical evidence show this to be realistic storage requirements.

A transposition table provides a flexible way of storing solution trees. While at any time entries from old (inferior) solution trees may be resident, they will be overwritten by newer entries when their space is needed. Garbage will collected incrementally. As long as the table is big enough to store the min or a max solution trees that are essential for the efficient operation of the algorithm, it provides for fast access and efficient storage. (Not to mention the move ordering and transposition benefits.)

*3.4 Starting Value*

Perhaps the biggest difference in the MTD algorithms is their first approximation of the minimax value: SSS* is optimistic, DUAL* is pessimistic and MTD-f is realistic. On average, there is a relationship between the starting bound and the size of the search trees generated. We have conducted tests that show that a sequence of MT searches to find the game value benefits from a start value close to the game value. Values like $+\infty$ or $-\infty$ as in SSS* and in DUAL* are in a sense the worst possible choices.

By doing minimal window searches, we hope to establish the left-most min solution tree and the left-most max solution tree as efficiently as possible. Doing searches with a different window can cause nodes to be expanded that are irrelevant for this proof. We have been unable to formulate this analytically, and instead present empirical evidence to support this conjecture.

Figure 10 validates the choice of a starting bound close to the game value. The figure shows the percentage of unique leaf evaluations of iterative deepening MTD-f. The data points are given as a percentage of the size of the search tree built by our best Alpha-Beta-searcher (Aspiration NegaScout). (Since iterative deepening algorithms are used, the cumulative leaf count over all previous depths is shown for the depths.) Given an initial guess of $h$ and the minimax value of $f$, the optimal search value for MTD-f, the graph plots the search effort expended for different values of $h - f$. To the left of the graph, MTD-f is closer to DUAL*, to the right it is closer to SSS*. To be better than Aspiration NegaScout, an algorithm must be less than the 100% baseline. A first guess close to $f$ makes MTD-f perform better than the 100% Aspiration NegaScout baseline. The guess must be close to $f$ for the effect to become significant. Thus, if MTD-f is to be effective, the $f$ obtained from the previous iteration must be a good indicator of the next iteration's value.

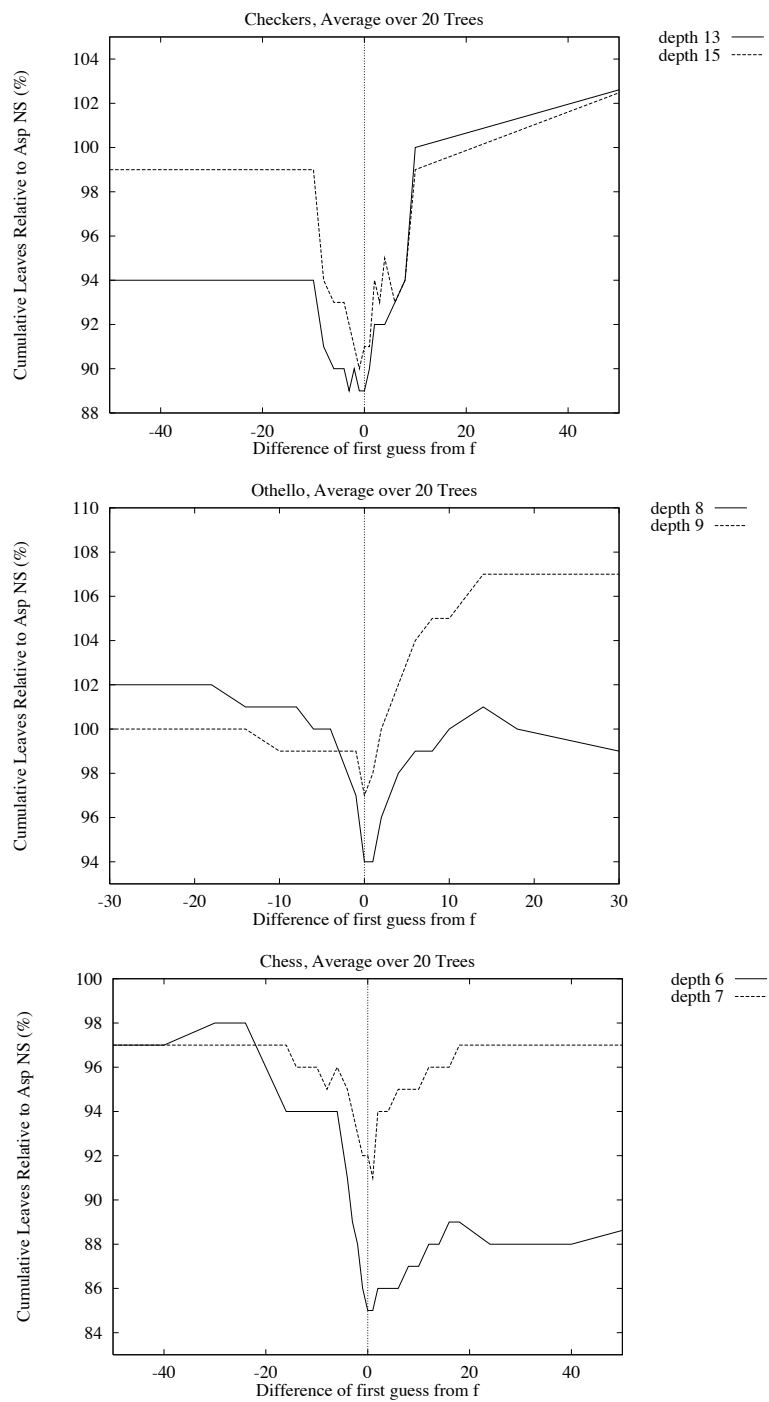

Figure 10: Tree size relative to the first guess $f$.

# 4  Experiments

There are three ways to evaluate a new algorithm: analysis, simulation or empirical testing. The emphasis in the literature has been on analysis and simulation. This is surprising given the large number of game-playing programs in existence.

## 4.1  Previous Work

The mathematical analysis of minimax search algorithms do a good job of increasing our understanding of the algorithms, but fail to give reliable predictions of performance (for example, [7, 16, 17]). The problem is that the game trees that are analyzed differ from the trees generated by real game-playing programs.

To overcome this deficiency, a number of authors have conducted simulations (for example, [8, 12, 13, 20]). In our opinion, the simulations did not capture the behavior of realistic search algorithms as they are used in game-playing programs. Instead, we decided to conduct experiments in a setting that was to be as realistic as possible. These experiments attempt to address the concerns we have with the simulation parameters:

- Variable branching factor: simulations use a fixed branching factor.

- High degree of ordering: most simulations have the quality of their move ordering below what is seen in real game-playing programs.

- Iterative deepening: simulations use fixed-depth searching. Game-playing programs use iterative deepening to seed memory (transposition table) with best moves to improve the move ordering. This adds overhead to the search, which is more than offset by the improved move ordering.

  Iterative deepening imposes a dynamic ordering on the tree. This means that different algorithms will search a tree differently. Thus, proofs that one fixed-depth algorithm searches less leaves than another fixed-depth algorithm (such as [26]), do not hold for iteratively-deepened searches.

- Memory: simulations assume either no storage of previously computed results, or unfairly bias their experiments by not giving all the algorithms the same storage. For iterative deepening to be effective, best move information from previous iterations must be saved in memory. In game-playing programs a transposition table is used. For best performance, this table may have to be of considerable size, since the move ordering information of the nodes of the previous search tree must be saved (which is exponential in the depth of the search).

- Tree size: simulations often use an inconsistent standard for counting leaf nodes. In conventional simulations (for example, [12]) each visit to a leaf node is counted for depth-first algorithms like NegaScout, whereas the leaf is counted only once for best-first algorithms like SSS* (because it was stored in memory, no re-expansion occurs). This problem is a result of the memory problem described above.

- Execution time: simulations are concerned with tree size, but practitioners are concerned with execution time. Simulations results do not necessarily correlate well with execution time. For example, there are many papers showing SSS* expands fewer leaf nodes than Alpha-Beta. However, SSS* implementations using Stockman's original algorithm have too much execution overhead to be competitive with Alpha-Beta.

- Value dependence: some simulations generate the value of a child independent of the value of the parent. However, there is usually a high correlation between the values of these two nodes in real games.

*4.2 Experiment Design*

To assess the feasibility of the proposed algorithms, a series of experiments was performed to compare Alpha-Beta, NegaScout, SSS* (MTD+∞), DUAL* (MTD−∞), MTD-f, and MTD-best.

Rather than use simulations, our data has been gathered from three game-playing programs: Chinook (checkers) [24], Keyano (Othello) and Phoenix (chess) [22]. All three programs are well-known in their respective domain. For our experiments, we used the original author's search algorithm which, presumably, has been highly tuned to the application. The only change we made was to disable search extensions and, in the case of Chinook and Phoenix, forward pruning. All programs used iterative deepening. The MTD algorithms would be repeatedly called with successively deeper search depths. All three programs used a standard transposition table with a maximum of $2^{20}$ entries. Testing showed that the solution trees could comfortably fit in tables of this size. For our experiments we used the original program author's transposition table data structures and code without modification.[1] At an interior node, the move suggested by the transposition table is always searched first (if known), and the remaining moves are ordered before being searched. Chinook and Phoenix use dynamic ordering based on the history heuristic [23], while Keyano uses static move ordering.

All three programs use transposition tables with only one bound (in contrast with our code in figure 1). Since MTD-bi and MTD-step manipulate both a max and a min solution tree, and therefore need to store both an upper and a lower bound at a node, we do not present these two algorithms. One of the points we are stressing in this paper is ease of implementation of the MT-framework. The amount of work involved in altering the information that is stored in the transposition tables would compromise this objective.

The MT code given in figure 1 has been expanded to include two details, both of which are common practice in game-playing programs. The first is a search depth parameter. This parameter is initialized to the depth of the search tree. As MT descends the search tree, the depth is decremented. Leaf nodes are at depth zero. The second is the saving of the best move at each node. When a node is revisited, the best move from the previous search is always considered first.

Conventional test sets in the literature proved to be poor predictors of performance. Positions in test sets are usually selected to test a particular characteristic or property of the game (such as tactical combinations in chess) and are not necessarily indicative of typical game conditions. For our experiments, the programs were tested using a set of 20 positions that corresponded to move sequences from tournament games. By selecting move sequences rather than isolated positions, we are attempting to create a test set that is representative of real game search properties (including positions with obvious moves, hard moves, positional moves, tactical moves, different game phases, etc.).

---

[1] As a matter of fact, since we implemented MT using null-window alpha-beta searches, we did not have to make any changes at all to the code other than the aforementioned disabling of forward pruning and search extensions. We only had to introduce the MTD driver code.

All three programs were run on 20 test positions, searching to a depth so that all searched roughly the same amount of time. Because of the low branching factor Chinook was able to search to depth 15, iterating two ply at a time. Keyano searched to 9 ply and Phoenix to 7, both one ply at a time.

*4.3 Base Line*

Many papers in the literature use Alpha-Beta as the base-line for comparing the performance of other algorithms (for example, [2, 10]). The implication is that this is the standard data point which everyone is trying to beat. However, game-playing programs have evolved beyond simple Alpha-Beta algorithms. Most use Alpha-Beta enhanced with minimal window search (usually PVS [2] or NegaScout [19]), iterative deepening, transposition tables, move ordering and an initial aspiration window. Since this is the typical search algorithm used in high-performance programs (such as Chinook, Phoenix and Keyano), it seems more reasonable to use this as our base-line standard.

The worse the base-line comparison algorithm chosen, the better other algorithms appear to be. By choosing NegaScout enhanced with aspiration searching (Aspiration NegaScout) as our performance metric, we are emphasizing that it is possible to do better than the "best" methods currently practiced and that, contrary to published simulation results, some methods are inferior.

Because we implemented the MTD algorithms (like SSS* and DUAL*) using MT (equivalent to null-window Alpha-Beta calls with a transposition table) we were able to compare a number of algorithms that were previously seen as very different. By using MT as a common proof-procedure, every algorithm benefited from the same enhancements concerning iterative deepening, transposition tables, and move ordering code. To our knowledge this is the first comparison of algorithms depth-first and best-first minimax search algorithms where all the algorithms are given identical resources.

*4.4 Experiment Results*

Figure 11 shows the performance of Chinook, Keyano and Phoenix, respectively, using the number of leaf evaluations (NBP or Number of Bottom Positions) as the performance metric. Figures 12 show the performance of these algorithms using the number of nodes in the search tree (interior and leaf) as the metric. The graphs show the cumulative number of nodes over all previous iterations for a certain depth (which is realistic since iterative deepening is used) relative to Aspiration NegaScout. Since we could not test MTD-bi and MTD-step properly without modifying the transposition table code, these results are not shown. The search depths reached by the programs vary greatly because of the differing branching factors. In checkers, the average branching factor is approximately 3 (there are typically 1.2 moves in a capture position; while roughly 8 in a non-capture position), in Othello it is 10 and in chess it is 36.

Over all three games, the best results are from MTD-f. Its leaf node counts are consistently better than Aspiration NegaScout, averaging at least a 5% improvement. More surprisingly is that MTD-f outperforms Aspiration NegaScout on the total node measure as well. Since each iteration requires repeated calls to MT (at least two and possibly many more), one might expect MTD-f to perform badly by this measure because of the repeated traversals of the tree. This suggests that MTD-f, on average, is calling MT close to the minimum number of times. For all three programs, MT gets called between 3 and 4 times on average. In contrast, the SSS* and DUAL* results are poor compared to NegaScout when all nodes in the search tree are considered. Each of these algorithms performs dozens and sometimes even hundreds of MT searches.

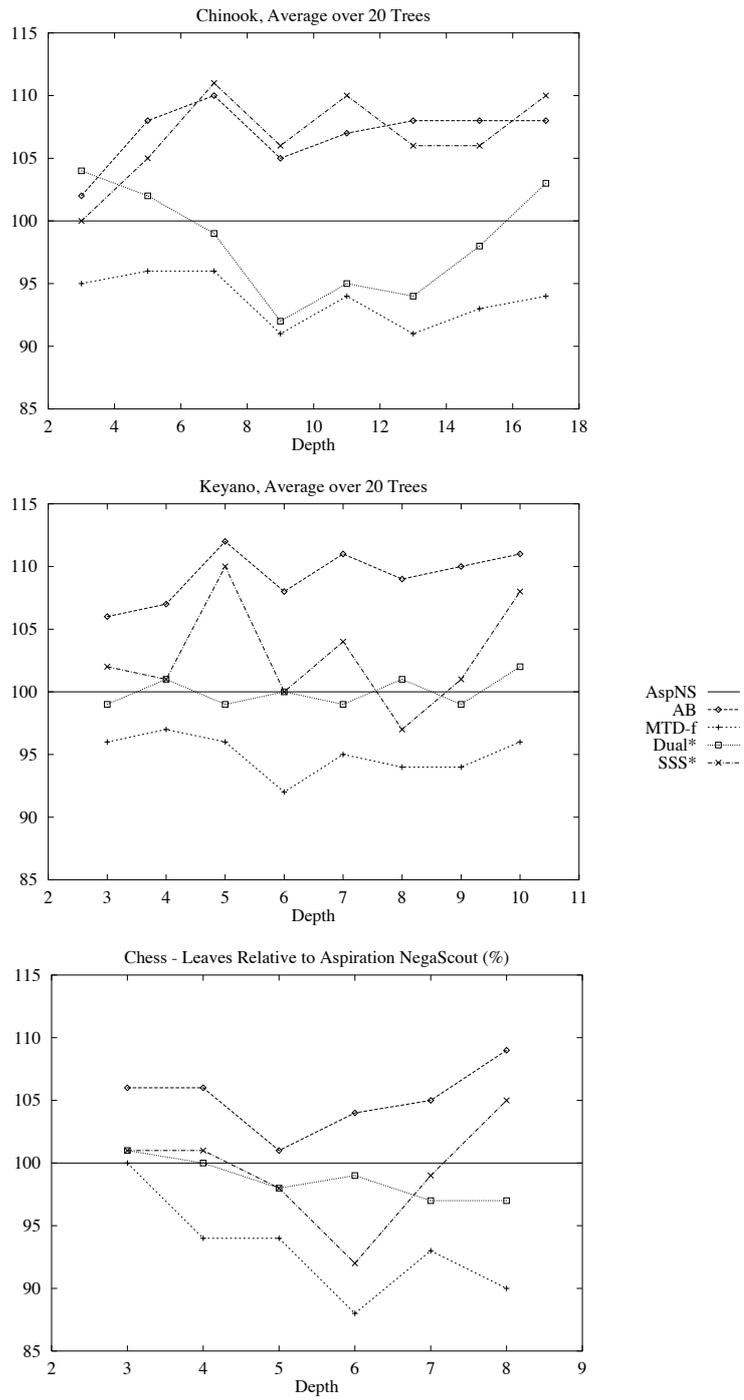

Figure 11: Leaf node count

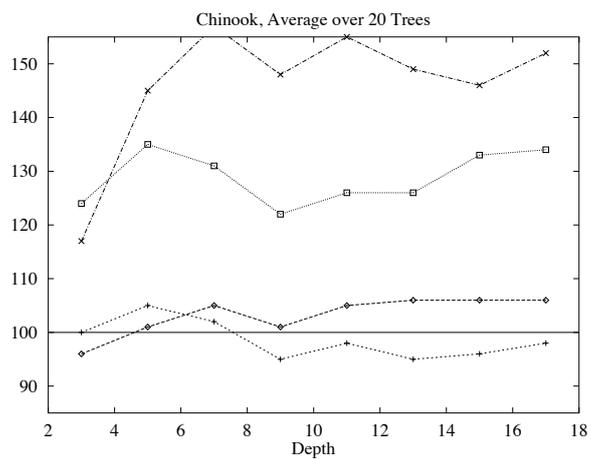

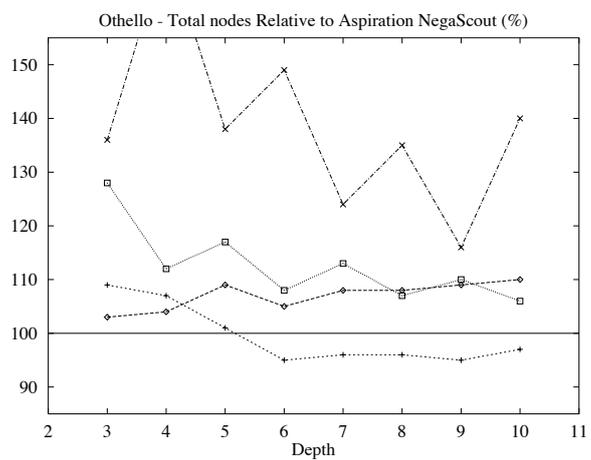

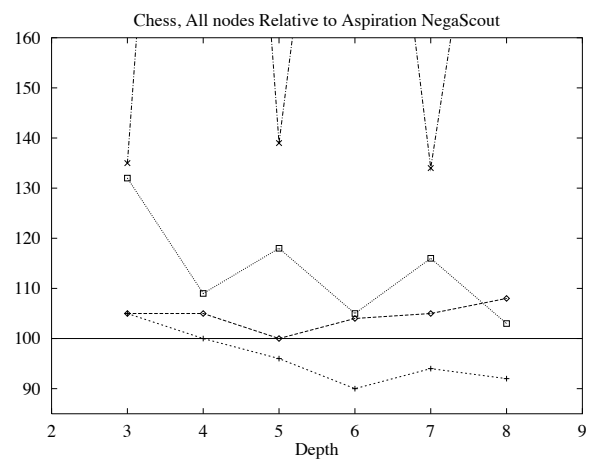

Figure 12: Total node count

The success of MTD-f implies that it is better to start the search with a good guess as to where the minimax value lies, rather than starting at one of the extremes ($+\infty$ or $-\infty$). Clearly, MTD-f out-performs Alpha-Beta, suggesting that binary-valued searches are more efficient than wide-windowed searches. Of more interest is the MTD-f and NegaScout comparison. If both searches have the correct best move from the previous iteration, then both find the value of that move and then build proof trees to show the remaining moves are inferior. Since NegaScout does this with a minimal window, it visits exactly the same leaf nodes that MT would. Since our results (and those from simulations) show NegaScout to be superior to Alpha-Beta, this suggests that limiting the use of wide-window searches is beneficial. MT takes this to the extreme by eliminating all wide-window searches.

What is the minimum amount of search effort that must be done to establish the minimax value of the search tree? If we know the value of the search tree is $f$, then two searches are required: MTD-f$(f - \varepsilon)$, which fails high establishing a lower bound on $f$, and MTD-f$(f + 1 - \varepsilon)$, which fails low and establishes an upper bound on $f$. Of the algorithms presented, MTD-f has the greatest likelihood of doing this minimum amount of work. The closer the approximation to $f$, the less the work that has to be done (see also figure 10). Considering this, it is not a surprise that both DUAL* and SSS* come our poorly. Their initial bounds for the minimax value are generally poor ($-\infty$ and $+\infty$ respectively), meaning that the many calls to MT result in significantly more interior nodes. In the literature, SSS* is regarded as good because it expands fewer leaf nodes than Alpha-Beta, and bad because of the algorithm complexity, storage requirements and storage maintenance overhead. Our results give the exact opposite view: SSS* is easy to implement because of the MT framework, uses as much storage as Aspiration NegaScout, and performs generally worse than Aspiration NegaScout when viewed in the context of iterative deepening and transposition tables. DUAL* is more efficient than SSS* but still comes out poorly in all the graphs measuring total node count. The Chinook leaf count graph is instructive, since here iterative deepening SSS* is generally *worse* than iterative deepening Alpha-Beta. This result runs opposite to both the proof and simulations for fixed depth SSS* and Alpha-Beta. An interesting observation is that the effectiveness of SSS* appears to be a function of the branching factor; the larger the branching factor, the better it performs.

Aspiration NegaScout is better than Alpha-Beta. This result is consistent with [23] which showed aspiration NegaScout to be a small improvement over Alpha-Beta when transposition tables and iterative deepening were used. NegaScout uses a wide-window search for the principal variation (PV) and all re-searches. The wide-window PV search result gives a good first approximation to the minimax value. That approximation is then used to search the rest of the tree with minimal window searches, which are equivalent to MT calls. If these refutation searches are successful (no re-search is needed), then NegaScout deviates from MTD-f only in the way it searches the PV for a value. MTD-f uses MT also for searching the PV, and for re-searches after a fail high. Since MTD-f does not derive its first guess from the tree like NegaScout does, it must get it externally. In our experiments the minimax value from the previous iterative deepening iteration was used for this purpose.

Simulation results show that for fixed depth searches, without transposition tables and iterative deepening, SSS*, DUAL* and NegaScout are major improvements over simple Alpha-Beta [8, 12, 20]. For example, one study shows SSS* and DUAL* building trees that are half the size of those built by Alpha-Beta [12]. This is in sharp contrast to the results reported here. Why is there such a disparity with the previously published work? Including transposition tables and iterative deepening in the experiment improves the search efficiency in two ways:

- improve move ordering so that the likelihood of the best move being considered first at a cutoff node is very high, and

- eliminate large portions of the search by having the search tree be treated as a search graph. A path that transposes into a path already searched can reuse the previously computed result.

The move ordering is improved to the extent that all algorithms are converging on the minimal search tree.

How effective is the move ordering? At a node with a cutoff, only one move should be examined. Data gathered from Phoenix, Keyano, and Chinook show an average of around 1.2, 1.2 and 1.1 move, respectively, are considered at cut nodes. On average, over 95% of the time the move ordering is successful. Clearly, the low ply numbers should have very good ordering because of the iterative deepening (typically 97-98%). Of more interest are these figures for the higher ply numbers. These nodes occur near the bottom of the tree and usually do not have the benefit of transposition table information to guide the selection of the first move to consider. Nevertheless, even these nodes have a success rate in excess of 90% (80% for Keyano, which does not have the history heuristic). Campbell and Marsland define *strongly ordered* tress to have the best move first 80% of the time [2]. Kaindl *et al.* called 90% *very strongly ordered* [8]. Our results suggest there should be a new category, *almost perfectly ordered*, corresponding to over 95% success of the first move at cutoff nodes. Given that the simulations used poorer move ordering than is seen in practice, it is not surprising that their results have little relation to ours.

### 4.5 Execution Time

The bottom line for practitioners is execution time. Since we did not have the resources to run all our experiments on identical and otherwise idle machines, we only show execution time graphs for ID MTD-f and ID Aspiration NegaScout (figure 13), the comparison that we think is the most interesting, since comparing results for the same machines we found that MTD-f is consistently the fastest algorithm[2]. It is about 5% faster than Aspiration NegaScout for checkers and Othello, and about 10% faster for chess, (depending in part on the quality of the score of the previous iteration).

---

[2]We only show the deeper searches, since the relatively fast shallower searches hamper accurate timings. The run shown is a typical example run on a Sun SPARC. We did experience different timings when running on different machines. It may well be that cache size plays an important role, and that tuning of the program can have a considerable impact.

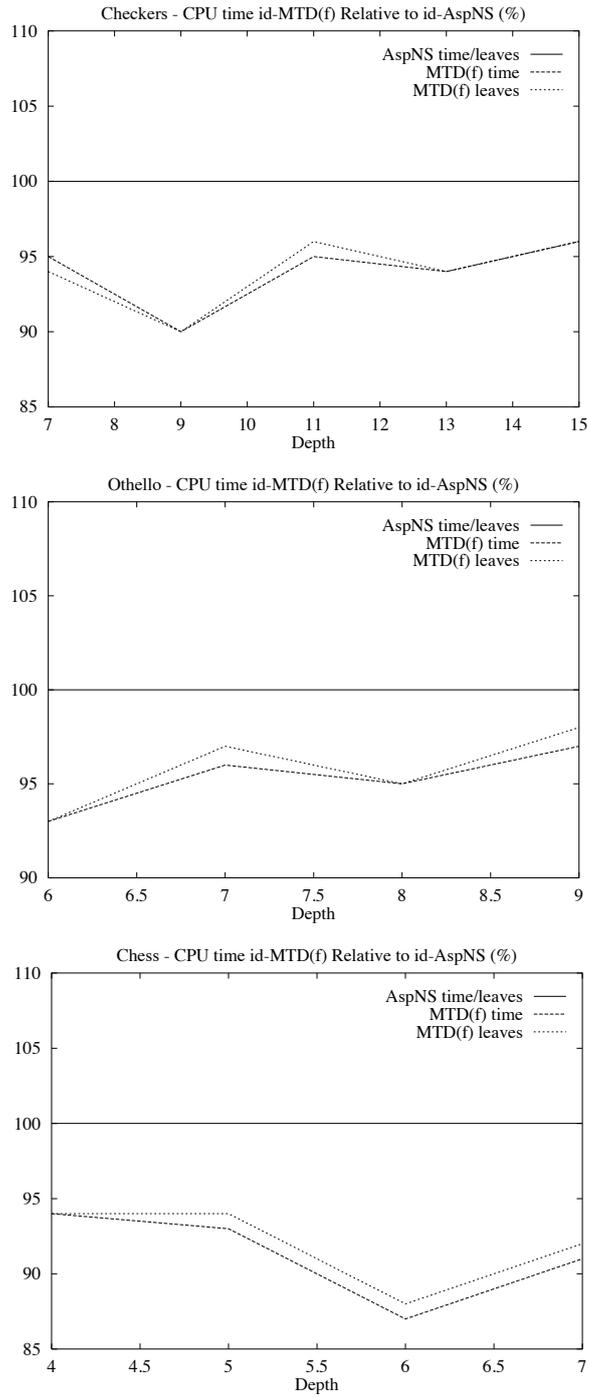

Figure 13: Leaf node count

*4.6 MTD-best*

The algorithms discussed so far are designed to find the minimax value of the tree. MTD-best is only interested in the best move and not the best value. Thus it has the potential to build even *smaller* trees. This algorithm uses the value of the previous iteration to try to establish only a lower bound on the best move, and then (if this was successful) an upper bound on the other moves. MTD-best proves the superiority of the best move by only constructing a lower bound. This algorithm saves on node expansions by not constructing an upper bound on the value of the best move. Figure 14 shows that MTD-best can be slightly better than MTD-f. This result must be taken in the context that MTD-best is trying to solve a simpler problem: find the best move, not the best value. If the choice of best move is the same as in the previous iteration, MTD-best is effective at proving that move is best. When the candidate best move turns out not to be best, MTD-best is inefficient since it ends up doing more work because its initial assumption (the best move) is wrong. Thus on a given search, MTD-best may be much better or much worse than MTD-f. In the test data, MTD-best was effective at proving the best move about half of the time. Since the algorithm is often performing a few percent better than MTD-f, the potential gains of proving a move best out-weigh the potential losses.

Our implementation of MTD-best is just a first attempt. We believe that there is still room for improving the algorithm.

## 5 Conclusions

Over thirty years of research have been devoted to improving the efficiency of alpha-beta searching. The MT family of algorithms are comparatively new, without the benefit of intense investigations. Given that MTD-f is already out-performing our best alpha-beta based implementations in real game-playing programs, future research can only make variations on this algorithm more attractive. MT is a simple and elegant paradigm for high performance game-tree search algorithms.

The purpose of a simulation is to exactly model an algorithm to gain insight into its performance. Simulations are usually performed when it is too difficult or too expensive to construct the proper experimental environment. The simulation parameters should be chosen to come as close to exactly modeling the desired scenario as is possible. In the case of game-tree searching, the case for simulations is weak. There is no need to do simulations when there are quality game-playing programs available for obtaining actual data. Further, as this paper has demonstrated, simulation parameters can be incorrect, resulting in large errors in the results that lead to misleading conclusions. In particular, the failure to include iterative deepening, transposition tables, and *almost perfectly ordered* trees in a simulation are serious omissions.

With easy to understand MT-based algorithms out-performing NegaScout, this leads to the obvious question: Why are you still using alpha-beta?

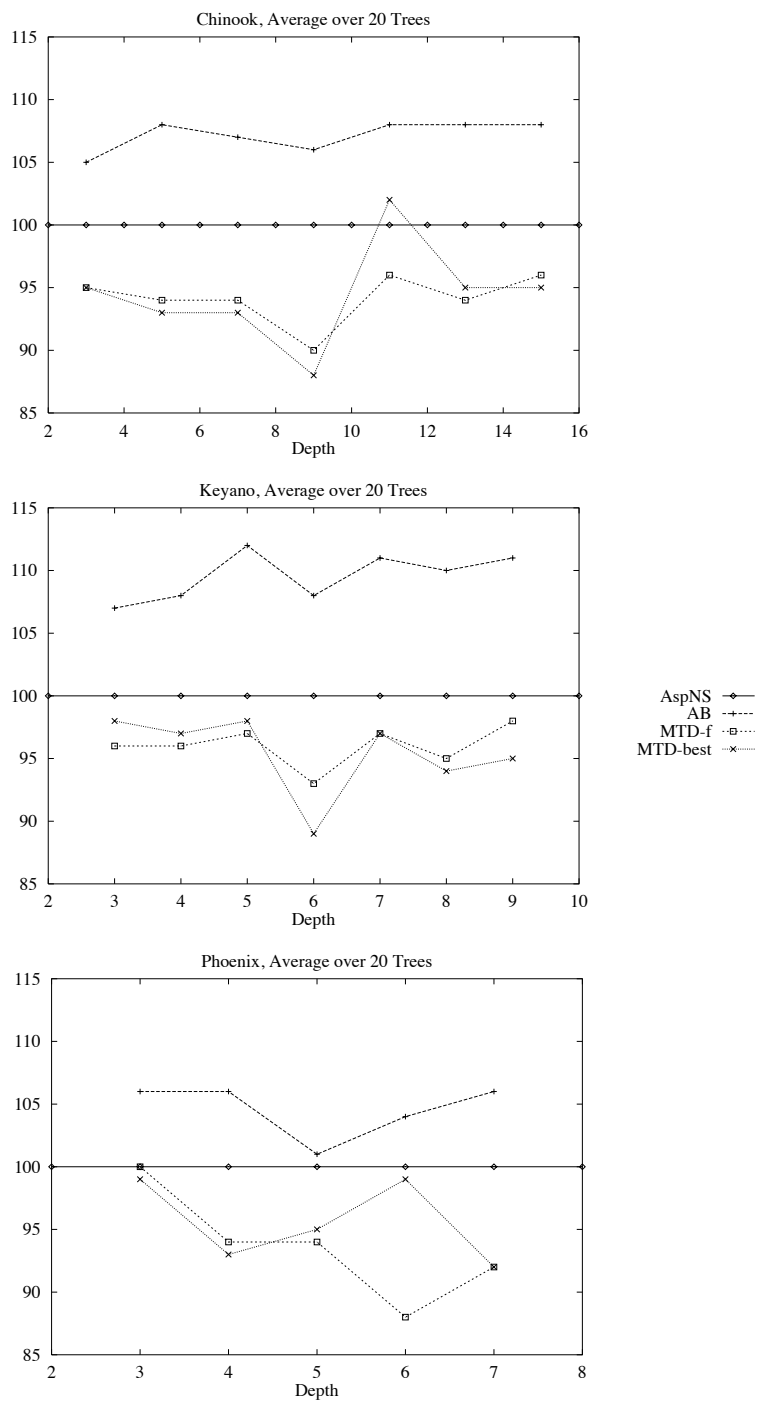

Figure 14: Leaf node count MTD-best

**Acknowledgements**

This work has benefited from discussions with Mark Brockington (author of Keyano), Yngvi Bjornsson and Andreas Junghanns. The financial support of the Netherlands Organization for Scientific Research (NWO), the Natural Sciences and Engineering Research Council of Canada (grant OGP-5183) and the University of Alberta Central Research Fund are gratefully acknowledged.
**References**

[1] Hans J. Berliner. The B* tree search algorithm: A best-first proof procedure. *Artificial Intelligence*, 12:23–40, 1979.

[2] Murray S. Campbell and T. Anthony Marsland. A comparison of minimax tree search algorithms. *Artificial Intelligence*, 20:347–367, 1983.

[3] K. Coplan. A special-purpose machine for an improved search algorithm for deep chess combinations. In M.R.B. Clarke, editor, *Advances in Computer Chess 3, April 1981*, pages 25–43. Pergamon Press, Oxford, 1982.

[4] Arie de Bruin, Wim Pijls, and Aske Plaat. Solution trees as a basis for game tree search. Technical Report EUR-CS-94-04, Department of Computer Science, Erasmus University Rotterdam, P.O. Box 1738, 3000 DR Rotterdam, The Netherlands, May 1994.

[5] Carl Ebeling. *All the Right Moves*. MIT Press, Cambridge, Massachusetts, 1987.

[6] Rainer Feldmann. *Spielbaumsuche mit massiv parallelen Systemen*. PhD thesis, Universität-Gesamthochschule Paderborn, May 1993.

[7] Toshihide Ibaraki. Generalization of alpha-beta and SSS* search procedures. *Artificial Intelligence*, 29:73–117, 1986.

[8] Hermann Kaindl, Reza Shams, and Helmut Horacek. Minimax search algorithms with and without aspiration windows. *IEEE Transactions on Pattern Analysis and Machine Intelligence*, PAMI-13(12):1225–1235, December 1991.

[9] Donald E. Knuth and Ronald W. Moore. An analysis of alpha-beta pruning. *Artificial Intelligence*, 6(4):293–326, 1975.

[10] T. Anthony Marsland. A review of game-tree pruning. *ICCA Journal*, 9(1):3–19, March 1986.

[11] T. Anthony Marsland and Alexander Reinefeld. Heuristic search in one and two player games. Technical report, University of Alberta, Paderborn Center for Parallel Computing, February 1993. Submitted for publication.

[12] T. Anthony Marsland, Alexander Reinefeld, and Jonathan Schaeffer. Low overhead alternatives to SSS*. *Artificial Intelligence*, 31:185–199, 1987.

[13] Agata Muszycka and Rajjan Shinghal. An empirical comparison of pruning strategies in game trees. *IEEE Transactions on Systems, Man and Cybernetics*, 15(3):389–399, May/June 1985.

[14] Judea Pearl. Asymptotical properties of minimax trees and game searching procedures. *Artificial Intelligence*, 14(2):113–138, 1980.

[15] Judea Pearl. The solution for the branching factor of the alpha-beta pruning algorithm and its optimality. *Communications of the ACM*, 25(8):559–564, August 1982.

[16] Judea Pearl. *Heuristics – Intelligent Search Strategies for Computer Problem Solving*. Addison-Wesley Publishing Co., Reading, MA, 1984.

[17] Wim Pijls and Arie de Bruin. Searching informed game trees. Technical Report EUR-CS-92-02, Erasmus University Rotterdam, Rotterdam, NL, October 1992. Extended abstract in Proceedings CSN 92, pp. 246–256, and Algorithms and Computation, ISAAC 92 (T. Ibaraki, ed), pp. 332–341, LNCS 650.

[18] Aske Plaat, Jonathan Schaeffer, Wim Pijls, and Arie de Bruin. Popular misconceptions about SSS* in practice. Technical Report TR-CS-94-17, Department of Computing Science, University of Alberta, Edmonton, AB, Canada, December 1994.

[19] Alexander Reinefeld. An improvement of the Scout tree-search algorithm. *ICCA Journal*, 6(4):4–14, 1983.

[20] Alexander Reinefeld. *Spielbaum Suchverfahren*. Volume Informatik-Fachberichte 200. Springer Verlag, 1989.

[21] Igor Roizen and Judea Pearl. A minimax algorithm better than alpha-beta? Yes and no. *Artificial Intelligence*, 21:199–230, 1983.

[22] Jonathan Schaeffer. *Experiments in Search and Knowledge*. PhD thesis, Department of Computing Science, University of Waterloo, Canada, 1986. Available as University of Alberta technical report TR86-12.

[23] Jonathan Schaeffer. The history heuristic and alpha-beta search enhancements in practice. *IEEE Transactions on Pattern Analysis and Machine Intelligence*, PAMI-11(1):1203–1212, November 1989.

[24] Jonathan Schaeffer, Joseph Culberson, Norman Treloar, Brent Knight, Paul Lu, and Duane Szafron. A world championship caliber checkers program. *Artificial Intelligence*, 53(2-3):273–290, 1992.

[25] D.J. Slate and L.R. Atkin. Chess 4.5 - the Northwestern University chess program. In P.W. Frey, editor, *Chess Skill in Man and Machine*, pages 82–118, New York, 1977. Springer-Verlag.

[26] George C. Stockman. A minimax algorithm better than alpha-beta? *Artificial Intelligence*, 12(2):179–196, 1979.

[27] Jean-Christophe Weill. The NegaC* search. *ICCA Journal*, 15(1):3–7, March 1992.